\documentclass[10pt]{article}

\usepackage[margin=1in]{geometry}
\usepackage{fontspec}
\usepackage{xcolor}
\usepackage{graphicx}
\usepackage{amsmath,amssymb}
\usepackage[hidelinks]{hyperref}

\usepackage{amsmath} 
\usepackage{amssymb} 
\usepackage{algorithm}
\usepackage{graphicx}
\usepackage{xcolor}
\usepackage{hyperref}
\usepackage{amsmath,amssymb}
\usepackage{wrapfig} 

\usepackage{booktabs}
\usepackage{caption}
\usepackage{cite}
\usepackage{float}
\floatstyle{plain}
\restylefloat{algorithm}

\usepackage{mdframed}
\usepackage{xspace}     
\usepackage{booktabs}
\usepackage[table]{xcolor}
\usepackage{tabularx}

\usepackage{algorithm}
\usepackage{algpseudocode}
\definecolor{AlgoBG}{HTML}{F7F8FA}
\algrenewcommand\algorithmicrequire{\textbf{Input:}}
\algrenewcommand\algorithmicensure{\textbf{Output:}}
\algrenewcommand\alglinenumber[1]{\scriptsize #1}
\algrenewcommand\algorithmicindent{1.0em}
\newcolumntype{Y}{>{\centering\arraybackslash}X}
\definecolor{TblHeader}{HTML}{F7F8FA}
\definecolor{TblSection}{HTML}{EEF2F6}
\definecolor{TblRule}{HTML}{D9DDE3}
\definecolor{ModernPink}{HTML}{FFE1E8}
\definecolor{modpink}{HTML}{E75480}
\newcommand{\modpink}[1]{\textcolor{modpink}{#1}}
\DeclareRobustCommand{\keyimg}{%
  \raisebox{-0.35ex}{\includegraphics[height=2.2ex]{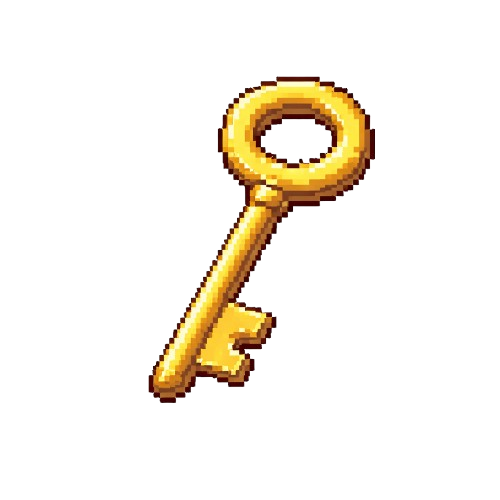}}%
}

\newcommand{\sectionband}[1]{%
  \addlinespace[2pt]
  \rowcolor{TblSection}\multicolumn{9}{@{}c@{}}{\textbf{\textcolor{black!70}{#1}}}\\
  \addlinespace[2pt]
}
\newcolumntype{L}{>{\raggedright\arraybackslash}X}
\newcolumntype{Y}{>{\centering\arraybackslash}X}  
\definecolor{Denim}{HTML}{1F4E79} 
\colorlet{tbmcolor}{Denim}
\newcommand{\tbm}[1]{\textbf{\textcolor{tbmcolor}{#1}}}
\newcommand{\tbmb}[1]{{\textcolor{tbmcolor}{#1}}}

\definecolor{poscolor}{RGB}{0,114,178}  
\definecolor{negcolor}{RGB}{213,94,0}   
\definecolor{keycolor}{RGB}{0,158,115}  
\definecolor{gencolor}{RGB}{0,0,0}

\newcommand{\xgen}{\ensuremath{\textcolor{gencolor}{\mathbf{x}}}}      
\newcommand{\ypos}{\ensuremath{\textcolor{poscolor}{\mathbf{y}^{+}}}} 
\newcommand{\yneg}{\ensuremath{\textcolor{negcolor}{\mathbf{y}^{-}}}} 
\newcommand{\kkey}{\ensuremath{\textcolor{keycolor}{\kappa}}}         
\newcommand{\dkey}{\ensuremath{\textcolor{keycolor}{d_{\kappa}}}}     
\newcommand{\Wpos}{\ensuremath{\textcolor{poscolor}{W^{+}}}}          
\newcommand{\Wneg}{\ensuremath{\textcolor{negcolor}{W^{-}}}}          
\newcommand{\mupos}{\ensuremath{\textcolor{poscolor}{\mu^{+}}}}        
\newcommand{\muneg}{\ensuremath{\textcolor{negcolor}{\mu^{-}}}}        

\usepackage{libertine}

\newcommand{\titlefont}{\libertineSB}
\newcommand{\headingfont}{\libertineSB}
\usepackage[libertine]{newtxmath}

\definecolor{abstractpink}{HTML}{FCEEF2}
\definecolor{rulegray}{HTML}{E6E0E3}
\definecolor{textgray}{HTML}{202020}

\hypersetup{
  pdftitle={Your Paper Title},
  colorlinks=true,
  urlcolor=blue,
  linkcolor=modpink,
  citecolor=modpink
}

\AtBeginDocument{\color{textgray}}
\newcommand{\uiucmark}{%
  \raisebox{0.25ex}{\includegraphics[height=1.5ex]{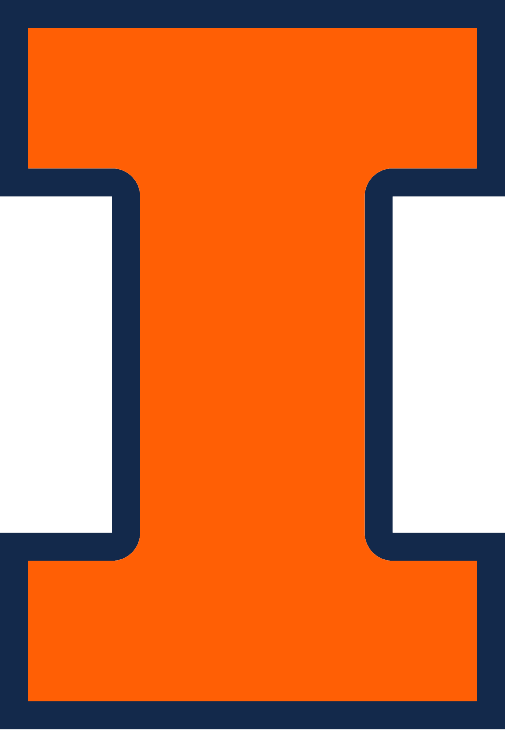}}%
}

\makeatletter
\renewcommand\section{\@startsection{section}{1}{\z@}%
  {1.8ex plus .2ex minus .2ex}%
  {0.8ex}%
  {\normalfont\Large\headingfont\color{textgray}}}
\renewcommand\subsection{\@startsection{subsection}{2}{\z@}%
  {1.5ex plus .2ex minus .2ex}%
  {0.6ex}%
  {\normalfont\large\headingfont\color{textgray}}}

\renewcommand{\maketitle}{%
  \begin{center}
    {\titlefont\fontsize{24}{30}\selectfont\color{textgray}\@title\par}
    \vspace{0.95em}

    \begin{minipage}[t]{0.46\textwidth}
      \centering
      {\normalsize\bfseries Gokul Puthumanaillam\,\uiucmark\par}
      \vspace{0.2em}
      {\small\ttfamily gokulp2@illinois.edu\par}
    \end{minipage}
    \hfill
    \begin{minipage}[t]{0.46\textwidth}
      \centering
      {\normalsize\bfseries Melkior Ornik\,\uiucmark\par}
      \vspace{0.2em}
      {\small\ttfamily mornik@illinois.edu\par}
    \end{minipage}

    \vspace{0.8em}
    {\small \uiucmark\ University of Illinois Urbana-Champaign\par}
  \end{center}
  \@thanks
  \vspace{1em}
  \noindent{\color{rulegray}\rule{\textwidth}{0.6pt}}\par
  \vspace{1.35em}
}
\makeatother

\newcommand{\styledabstract}[1]{%
  \begin{center}
    \begingroup
    \setlength{\fboxsep}{15pt}%
    \colorbox{abstractpink}{%
      \parbox{0.93\linewidth}{%
        \color{textgray}%
        {\headingfont\large Abstract}\par\medskip
        #1%
      }%
    }%
    \endgroup
  \end{center}
  \vspace{1.0em}
}

\title{\keyimg\ Amortizing Trajectory Diffusion with Keyed Drift Fields}
\author{}
\date{}

\begin{document}

\setlength{\columnsep}{0.18in}
\emergencystretch=1em

\twocolumn[{%
\maketitle

\styledabstract{%
Diffusion-based trajectory planners can synthesize rich, multimodal action sequences for offline reinforcement learning, but their iterative denoising incurs substantial inference-time cost, making closed-loop planning slow under tight compute budgets. 
We study the problem of achieving diffusion-like trajectory planning behavior with one-step inference, while retaining the ability to sample diverse candidate plans and condition on the current state in a receding-horizon control loop. Our key observation is that conditional trajectory generation fails under naïve distribution-matching objectives when the similarity measure used to align generated trajectories with the dataset is dominated by unconstrained future dimensions. In practice, this causes attraction toward average trajectories, collapses action diversity, and yields near-static behavior.
Our key insight is that conditional generative planning requires a conditioning-aware notion of neighborhood: trajectory updates should be computed using distances in a compact key space that reflects the condition, while still applying updates in the full trajectory space. Building on this, we introduce\keyimg \modpink{Keyed Drifting Policies} (KDP), a one-step trajectory generator trained with a drift-field objective that attracts generated trajectories toward condition-matched dataset windows and repels them from nearby generated samples, using a stop-gradient drifted target to amortize iterative refinement into training. At inference, the resulting policy produces a full trajectory window in a single forward pass.
Across standard RL benchmarks and real-time hardware deployments, KDP achieves strong performance with one-step inference and substantially lower planning latency than diffusion sampling.
Project website, code and videos: \modpink{\hyperlink{https://keyed-drifting.github.io/}{https://keyed-drifting.github.io/}}
}

\vspace{0.8em}
}]

\section{INTRODUCTION}
Diffusion-based trajectory models have emerged as a compelling foundation for robot decision-making and offline reinforcement learning (RL) because they can model complex, multimodal distributions over future behavior directly from data \cite{janner_planning_2022,ajay_is_2023,chi_diffusion_2024}. 
In a typical deployment, a diffusion model is used inside a receding-horizon control loop: at each control step, the system conditions on the current observation and optionally a goal or desired return, samples a short trajectory segment, executes the first action, and replans \cite{ajay_is_2023,zhou_diffusion_2025,chi_diffusion_2024}. 
Despite their flexibility, diffusion planners are often impractical at high control rates because they require many sequential denoising steps, leading to substantial latency and large numbers of neural network evaluations \cite{ho_denoising_2020,song_denoising_2022,zhou_diffusion_2025}. 
This paper asks a concrete question: how can we obtain diffusion-like trajectory planning behavior with one-step inference while preserving conditional controllability and sample diversity in a receding-horizon control loop?

Prior work reduces diffusion-planning latency either by cutting denoising steps \cite{song_denoising_2022,cac,salimans_progressive_2022,song_consistency_nodate,luo_latent_2023} or by iteratively sampling and selecting/guiding candidates with various signals \cite{lu_contrastive_2023,chen_offline_2023-1}. Both approaches remain tied to an iterative sampling loop, requiring added tuning as compute budgets tighten.
We take a different route: amortize refinement into training to obtain a one-step conditional trajectory generator usable for planning. Na\"ively matching the data distribution in full window space is brittle---in high-dimensional state-action sequences, distances are dominated by unconstrained future components, pulling samples toward averages, collapsing action diversity, and yielding inert closed-loop behavior.

Our key observation is that \tbmb{conditional trajectory generation requires a conditioning-aware notion of neighborhood}: for a given condition, the learning signal should preferentially compare against data trajectories that match that condition, rather than against arbitrary trajectories that are close under a high-dimensional, poorly aligned metric. 

\begin{figure}
    \centering
    \includegraphics[width=\linewidth, trim={0 9pt 0 0}, clip]{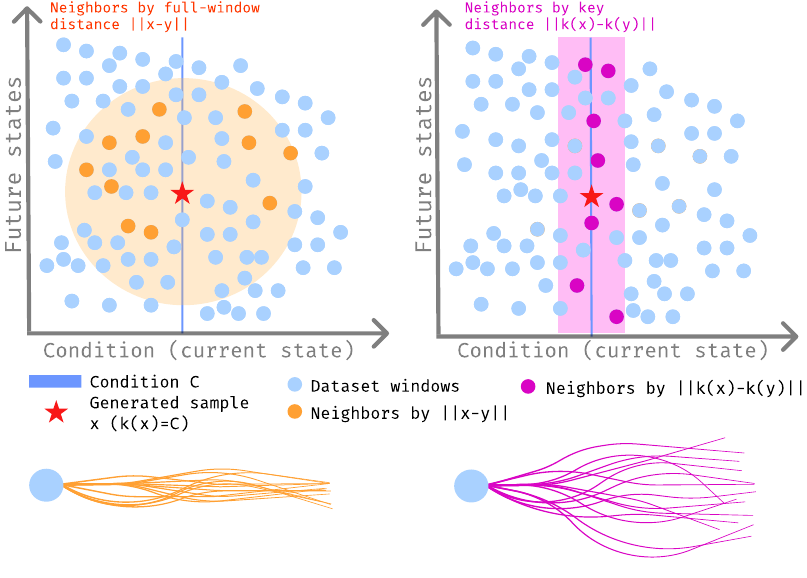}
    \caption{
    Given condition \(c\), full-window distance selects neighbors dominated by unconstrained future dimensions (orange), encouraging mode-averaged trajectories (left bottom). Key distance selects condition-matched neighbors (magenta), preserving diverse conditional trajectories (right bottom).}
    \label{fig:placeholder}
\end{figure}

We operationalize this observation with \keyimg{\modpink{Keyed Drifting Policies (KDP)}}, a one-step trajectory generator trained with a keyed drift field objective~\cite{deng2026generativemodelingdrifting} that pulls samples toward condition-matched dataset windows and repels them from nearby generated samples via a stop-gradient drifted target, amortizing refinement into the model parameters. KDP is deployed in the same receding-horizon planning interface as diffusion planners, and we contribute: (i) a practical keyed drift-field formulation for conditional trajectory generation that avoids collapse under hard conditioning constraints; (ii) a one-step planning recipe that preserves the diffusion-planner interface while eliminating the \(T\)-step sequential denoising loop; and (iii) an end-to-end evaluation spanning standard simulation benchmarks and real-time closed-loop hardware deployments under strict latency budgets.

\section{Related Work}
Our work sits at the intersection of diffusion-based planning, offline RL, and distribution matching. 

\noindent \tbm{Offline RL:}
Offline RL learns policies from fixed datasets \cite{bc} and combats distribution shift via behavior constraints, regularization, and conservative value learning \cite{cql,iql}. Practical variants blend cloning-style objectives with critic learning or advantage weighting \cite{fujimoto_minimalist_2021,wang_critic_2020}, while model-based offline RL uses learned dynamics with pessimism to mitigate compounding error \cite{mopo, morel, mbop}. Sequence-model approaches reframe decision making as conditional sequence prediction \cite{chen_decision_2021,janner_offline_2021,brandfonbrener_when_nodate}. These families are effective but usually yield reactive policies or require costly decoding or optimization, and they do not directly address reducing inference-time iterations for sampling-based planning.

\noindent \tbm{Diffusion-based planners:}
Diffusion-based planners model future trajectories or actions and deploy them by sampling candidate plans or actions \cite{janner_planning_2022,ajay_is_2023,dql}. They support flexible conditioning via inpainting, return conditioning, or guidance \cite{janner_planning_2022,ajay_is_2023} and have been extended through hierarchical, latent, and energy-guided variants \cite{li_hierarchical_2023,li_efficient_2023}. Robotics variants apply diffusion to trajectory generation and control under constraints \cite{zhou_diffusion_2025,chi_diffusion_2024}. Despite strong modeling capacity, these methods typically rely on iterative denoising, which increases inference latency and can introduce closed-loop brittleness for real-time planning.

\noindent \tbm{One-step generators:}
Faster samplers reduce denoising steps via deterministic trajectories, improved parameterizations, or specialized solvers, and guidance mechanisms steer samples toward conditions without explicit classifiers \cite{song_denoising_2022,cac}. Distillation and alternative objectives target few-step generation and reduces per-step compute \cite{salimans_progressive_2022,song_consistency_nodate,luo_latent_2023}. Score-based foundations formalize diffusion and score learning in continuous time and in latent spaces, and hybrid objectives combine diffusion training with GAN-style efficiency \cite{song_score-based_2021,vahdat_score-based_2021,jeon_spi-gan_2024}. These approaches often require additional stages or tuning; our focus is amortizing refinement into training for one-step conditional planning.

\noindent \tbm{Distribution matching:}
Kernel and particle-based perspectives emphasize that distribution matching depends on the neighborhood metric: mean-shift-style updates and density-gradient estimators motivate attraction dynamics, while kernel mean embeddings formalize distributional discrepancies \cite{fukunaga_estimation_1975,comaniciu_mean_2002,muandet_kernel_2017}. Recent methods combine attraction with kernel-based repulsion to maintain diversity, including policy-gradient adaptations \cite{liu_stein_2016,liu_stein_2017}. Energy-based and score-based learning provide complementary views: score matching links learning to density-gradient estimation, while energy-based learning, and noise-contrastive estimation use negative samples to fit unnormalized models \cite{hyvarinen_estimation_nodate,gutmann_noise-contrastive_nodate}. Adversarial objectives and contrastive learning operationalize attraction/ repulsion through learned discrepancies and positive/negative sampling \cite{oord_representation_2019,chen_simple_2020}. Autoregressive sequence models provide another form of distribution matching by fitting trajectory distributions directly through conditional sequence prediction \cite{chen_decision_2021,janner_offline_2021}.

\section{Background}

\subsection{Trajectory diffusion models for planning}
We consider diffusion models that generate short {trajectory windows} for receding-horizon control. Let a window be represented as $x_0 \in \mathbb{R}^{H \times D} (D = d_s + d_a)$, where each timestep concatenates state and action, i.e., $x_0[t] = (s_t, a_t)$. At decision time, the model is conditioned on a context $c$ derived from the current observation (and optionally a goal). In this paper, the conditioning signal we primarily care about is the current state, written as $c = s_0$.

Diffusion introduces a sequence of latent variables $\{x_t\}_{t=1}^T$ via a forward noising process \cite{ho_denoising_2020}
\begin{equation}
q(x_t \mid x_{t-1}) = \mathcal{N}\!\left(\sqrt{\alpha_t}\, x_{t-1},\; (1-\alpha_t) I\right),
\end{equation}
with $\alpha_t = 1 - \beta_t$ and a user-chosen noise schedule $\{\beta_t\}_{t=1}^T$. Defining $\bar{\alpha}_t = \prod_{i=1}^t \alpha_i$, the marginal admits the closed form
\begin{equation}
q(x_t \mid x_0) = \mathcal{N}\!\left(\sqrt{\bar{\alpha}_t}\, x_0,\; (1-\bar{\alpha}_t) I\right),
\end{equation}
also written as $x_t = \sqrt{\bar{\alpha}_t}\,x_0 + \sqrt{1-\bar{\alpha}_t}\,\varepsilon$ with $\varepsilon \sim \mathcal{N}(0,I)$.

A standard parameterization trains a denoiser $\varepsilon_\theta$ to predict the injected noise under conditioning $c$:
\begin{equation}
\min_\theta\; \mathbb{E}_{x_0 \sim \mathcal{D},\, t \sim \{1,\dots,T\},\, \varepsilon \sim \mathcal{N}(0,I)}
\left[\left\lVert \varepsilon - \varepsilon_\theta(x_t, t, c)\right\rVert_2^2\right].
\end{equation}
At inference, trajectories are sampled by reversing the process from $x_T \sim \mathcal{N}(0,I)$ using a learned reverse transition
\begin{equation}
p_\theta(x_{t-1} \mid x_t, c) = \mathcal{N}\!\left(\mu_\theta(x_t,t,c),\; \sigma_t^2 I\right),
\end{equation}
where the mean typically takes the DDPM \cite{ho_denoising_2020} form 
\begin{equation}
\mu_\theta(x_t,t,c)
=
\frac{1}{\sqrt{\alpha_t}}
\left(
x_t
-
\frac{1-\alpha_t}{\sqrt{1-\bar{\alpha}_t}}\;\varepsilon_\theta(x_t,t,c)
\right),
\end{equation}
and $\sigma_t^2$ is set from the schedule.

\subsection{Receding-horizon control and conditioning}
In receding-horizon deployment, at each environment step we observe the current state $s$ and set $c=s$. A sampled window $\hat{x}_0$ yields the control action by extracting its first action component: $a = \hat{x}_0[0,\, d_s{:}d_s{+}d_a]$.
Implementations \cite{janner_planning_2022} further improve reliability by sampling $K$ candidate windows $\{\hat{x}_0^{(k)}\}_{k=1}^K$ and selecting via a learned scorer $J_\phi$.
\subsection{Computational challenges with diffusion planning}
Diffusion planning is dominated by {sequential} denoising: generating one candidate trajectory requires \(T\) denoiser evaluations, and best-of-\(K\) selection scales this to \(\approx K T\) evaluations per control step, plus scoring. Decreasing \(T\) reduces latency but changes the sampling distribution and harms conditional fidelity; increasing \(K\) improves selection but increases cost linearly.
A natural alternative is a one-step conditional generator \(g_\psi(z,c)\) that outputs a full horizon-\(H\) window in a single forward pass. However, naïvely training such a generator by matching distributions in full window space is unstable: distances and losses are dominated by {unconstrained} future components, encouraging collapse toward average trajectories and producing inert controllers despite numerically stable training. 

\section{\texorpdfstring{\keyimg\ Keyed Drifting Policies}{Keyed Drifting Policies}}
\begin{figure*}
    \centering
    \includegraphics[width=0.9\linewidth]{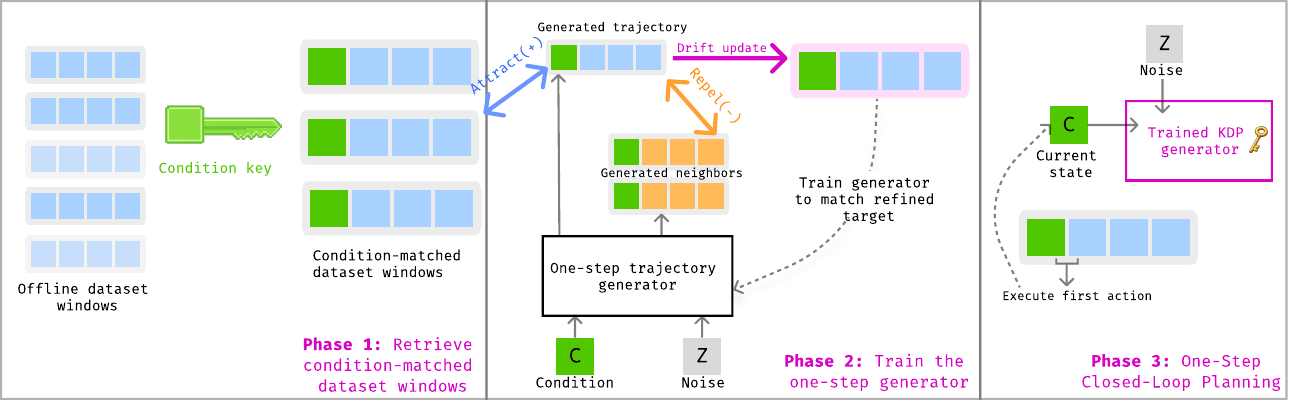}
    \caption{Overview of KDP. A condition key retrieves condition-matched dataset windows, which train a one-step generator by attracting trajectories toward matched data and repelling them from nearby generated samples, enabling one-step closed-loop planning at inference.}
    \label{fig:placeholder}
\end{figure*}

\label{sec:method}

We introduce \keyimg{\modpink{Keyed Drifting Policies (KDP)}}, which train a conditional generator by evolving its training-time pushforward distribution with a drifting field, following the {stop-gradient drifted target} recipe of drifting models~\cite{deng2026generativemodelingdrifting}.
The modification for conditional planning is that we make neighborhood structure \emph{conditioning-aware}: when we decide which dataset windows should influence a generated sample, we measure similarity in a \emph{key space} aligned with the condition, while applying updates in the full trajectory space.

\subsection{Trajectory windows and hard conditioning constraints}
\label{sec:method_windows}
Recall that a trajectory window is represented as \(x\) and $x[t] = (s_t, a_t)$.
We use the block operators:
$\mathrm{state}(x)[t] = x[t,0{:}d_s],$
$\mathrm{act}(x)[t] = x[t,d_s{:}D].$

In receding-horizon control, the condition is the current observation (state conditioning),
\(c=s_0 \in \mathbb{R}^{d_s}\).
Because the plan must start at the true current state, we enforce this as a hard constraint via a clamping operator \(\mathcal{C}_c(\cdot)\):
\begin{equation}
\mathcal{C}_{c}(x)[0,0{:}d_s] = c,
\qquad
\mathcal{C}_{c}(x)[t,:] = x[t,:]\;\; \text{otherwise.}
\end{equation}
Clamping is necessary because small prediction errors in the conditioned dimensions accumulate across replanning steps, which can destabilize the closed-loop behavior.

\paragraph{Goal-conditioned inpainting}
For goal-conditioned tasks we augment the condition with a goal vector \(g\in\mathbb{R}^{d_g}\) and
enforce goal satisfaction by inpainting selected state coordinates at a designated timestep \(t_g\).
Let \(\mathcal{I}\subseteq \{1,\dots,d_s\}\) denote the goal-relevant state indices. We define an
extended clamping operator \(\mathcal{C}_{(c,g)}\) that overwrites both the initial state and the
goal coordinates:
\begin{equation}
\mathcal{C}_{(c,g)}(x)[0,d_a{:}D] = c,
\qquad
\mathcal{C}_{(c,g)}(x)[t_g, d_a+\mathcal{I}] = g,
\end{equation}
leaving all other entries unchanged. During training we apply the same mask to the drift field so
drift updates cannot modify clamped coordinates.

\subsection{Conditional pushforward at training}
\label{sec:method_pushforward}
Let \(g_\psi\) be a conditional generator that maps noise \(z \sim \mathcal{N}(0,I)\) and condition \(c\) to a trajectory window:
\begin{equation}
\xgen = \mathcal{C}_{c}\!\left(g_\psi(z,c)\right) \in \mathbb{R}^{H \times D}.
\end{equation}
For a fixed condition \(c\), the model induces a conditional pushforward distribution
\(q_\psi(\cdot \mid c)\).
During stochastic optimization, the parameter sequence \(\{\psi_i\}\) induces a sequence of conditional pushforwards \(\{q_{\psi_i}(\cdot \mid c)\}\).
Drifting models~\cite{deng2026generativemodelingdrifting} view training as an evolution of the pushforward distribution governed by a drifting field; we adopt the same viewpoint, but the conditional setting forces a question that does not arise in unconditional generation: \emph{how should we define neighborhood structure so that the learning signal reflects the condition rather than being dominated by unconstrained future dimensions?}

Empirically, if similarity is measured over full windows early in training, the unconstrained future dominates distance computations and leads to unstable or collapsed behavior (e.g., action variance collapse and near-static rollouts).
KDP addresses this by defining neighborhoods in a key space aligned with the condition.

\subsection{Key space and conditioning-aware neighborhoods}
\label{sec:method_key}
KDP introduces a \textit{key map} \(\kkey(\cdot)\) used \emph{only} for neighborhood formation.
In the state-conditioned case we use the initial state block: $\kkey(x) = \mathrm{state}(x)[0] \in \mathbb{R}^{d_s}.$

Key-space distances are Euclidean:
\begin{equation}
\dkey(x,y) = \left\lVert \kkey(x) - \kkey(y) \right\rVert_2.    
\end{equation}
Because we always clamp \(\mathrm{state}(\xgen)[0]=c\), the key of a generated window is exactly the condition.
As a result, neighborhood selection depends on \(c\) rather than on the current (potentially noisy) future portion of \(\xgen\).
This is the mechanism by which KDP makes conditional learning signals robust in high-dimensional trajectory spaces.

\subsection{A keyed drifting field in full trajectory space}
\label{sec:method_drift}
Having fixed \emph{how} we form neighborhoods (Sec.~\ref{sec:method_key}), we still need a \emph{training-time refinement rule}: given a clamped generated window \(\xgen_i\), what update direction should move it toward the conditional data distribution while avoiding mode collapse?
Following drifting models~\cite{deng2026generativemodelingdrifting}, we construct a {drifting field} \(V(\xgen)\) with two roles:
(i) \textcolor{poscolor}{\emph{attraction}} toward dataset windows that match the condition, and
(ii) \textcolor{negcolor}{\emph{repulsion}} away from nearby generated windows to maintain diversity.
KDP uses the \textcolor{keycolor}{key distance} \(\dkey\) to decide which windows are near, but applies the update in the full window space \(\mathbb{R}^{H\times D}\).

\noindent \tbm{From mean-shift to a keyed drift.}
A convenient way to think about drift is as a mean-shift-like update.
Given a normalized kernel \(\tilde{k}(\cdot,\cdot)\) and sources of samples: \(\ypos \sim p\) from the dataset and \(\yneg \sim q\) from the current model---a standard attraction--repulsion field can be written as
\begin{equation}
\label{eq:keyed_drift_general}
\text{\footnotesize$\displaystyle
V_{p,q}(\xgen)
=
\mathbb{E}_{\ypos \sim p}\!\big[\,\tilde{k}_{\kkey}(\xgen,\ypos)\,\ypos\,\big]
\;-\;
\mathbb{E}_{\yneg \sim q}\!\big[\,\tilde{k}_{\kkey}(\xgen,\yneg)\,\yneg\,\big],
$}
\end{equation}
where \(\tilde{k}_{\kkey}\) emphasizes that similarity is computed in \textcolor{keycolor}{key space}.
This form is equivalent to the mean of differences presentation in drifting models~\cite{deng2026generativemodelingdrifting} (the \(\xgen\) terms cancel between attraction and repulsion), and it makes explicit what KDP needs: a way to compute \(\tilde{k}_{\kkey}\) that reflects the condition.

\noindent \tbm{Mini-batch construction.}
In stochastic training, we approximate the expectations in Eq.~\eqref{eq:keyed_drift_general} with minibatch averages.
Let \(\{\ypos_j\}_{j=1}^B\) be a minibatch of dataset windows.
We set the condition for the \(i\)-th sample to be the dataset key, i.e., 
\begin{equation}
c_i = \kkey(\ypos_i);\text{ }
\xgen_i = \mathcal{C}_{c_i}\!\left(g_\psi(z_i,c_i)\right);\text{ }
 z_i \sim \mathcal{N}(0,I),
\end{equation}
so that the empirical distribution of conditions seen by the generator matches the data.
We use the current minibatch of generated samples as negatives,
\(\yneg_j := \xgen_j\),
so repulsion is always defined \emph{relative to the model's current conditional output distribution}.

\noindent \tbm{Keyed kernel weights.}
A drifting field is only as useful as its notion of nearby.
If we computed similarity in full window space early in training, distances would be dominated by unconstrained future components, and the resulting weights would not reflect the condition.
KDP therefore computes kernel weights using key distances:
\begin{equation}
D^+_{ij} = \dkey(\xgen_i,\ypos_j),
\qquad
D^-_{ij} = \dkey(\xgen_i,\yneg_j).
\end{equation}
We adopt the same exponential kernel family as~\cite{deng2026generativemodelingdrifting} implemented via a softmax normalization:
\begin{equation}
\label{eq:keyed_softmax_weights}
\begin{aligned}
\Wpos_{ij}
&=
\frac{\exp\!\big(-D^+_{ij}/\tau\big)}
{\sum_{k=1}^B \exp\!\big(-D^+_{ik}/\tau\big)}, \\
\Wneg_{ij}
&=
\frac{\exp\!\big(-D^-_{ij}/\tau\big)}
{\sum_{k=1}^B \exp\!\big(-D^-_{ik}/\tau\big)} .
\end{aligned}
\end{equation}
Crucially, because \(\xgen_i\) is clamped, \(\kkey(\xgen_i)\) is determined by \(c_i\); thus \(D^+_{ij}\) compares {conditions} rather than unconstrained futures, and \(\Wpos\) concentrates on condition-matched windows.

\noindent \tbm{Self-negative masking.}
If we include \(j=i\) in the repulsion softmax, then \(D^-_{ii}=0\) makes \(\Wneg_{ii}\) dominate, yielding \(\muneg_i \approx \xgen_i\).
This collapses the repulsion term into a no-op and reduces the drift to a pure attraction update, which empirically encourages mode averaging.
We therefore exclude self-negatives by masking the diagonal before normalization (equivalently \(D^-_{ii}\leftarrow +\infty\) in Eq.~\eqref{eq:keyed_softmax_weights}), matching the self-negative masking fix used in our implementation.

\noindent \tbm{Attraction and repulsion means in full window space.}
We form weighted averages of {full} windows:
\begin{equation}
\label{eq:keyed_mus}
\mupos_i = \sum_{j=1}^B \Wpos_{ij}\,\ypos_j,
\qquad
\muneg_i = \sum_{j=1}^B \Wneg_{ij}\,\yneg_j.
\end{equation}
Here \(\mupos_i\) is a condition-local prototype of the dataset in full trajectory space, \(\muneg_i\) is the corresponding prototype of the model's current samples under the same key-locality notion.
The drift direction is their difference: $V_i^{(\tau)} = \mupos_i - \muneg_i.$
This is precisely the batch approximation of Eq.~\eqref{eq:keyed_drift_general}.

\noindent \tbm{Multi-temperature averaging.}
The temperature \(\tau\) controls how local the neighborhood is: smaller \(\tau\) emphasizes nearest neighbors in key space, while larger \(\tau\) averages over a broader set of condition-similar windows.
To reduce sensitivity to a single bandwidth choice, we average drifts across a small set of temperatures 
\begin{equation}
 \{\tau_m\}_{m=1}^M: V_i = \frac{1}{M}\sum_{m=1}^M V_i^{(\tau_m)}.
\end{equation}

\noindent \tbm{Constraint-respecting drift (masking) and stable magnitudes (normalization).}
Because KDP enforces hard constraints via \(\mathcal{C}\), refinement must not alter clamped entries.
We therefore apply a binary mask \(M \in \{0,1\}^{H \times D}\) that zeros clamped coordinates: 
\begin{equation}
    V_i \leftarrow M \odot V_i.
\end{equation}

Finally, we normalize nonzero drift vectors per sample: 
\begin{equation}
V_i \leftarrow {V_i}/\big({\sqrt{\frac{1}{HD}\lVert V_i \rVert_2^2} + \epsilon}\big).
\end{equation}
This keeps equilibrium at \(V_i=0\) unchanged and prevents drift from varying wildly.

\subsection{Amortizing refinement with a stop gradient drifted target}
\label{sec:method_objective}
The keyed drift field \(V_i\) defines a refinement step that would improve \(\xgen_i\) if applied iteratively.
Instead, we amortize refinement into \(g_\psi\) using a stop-gradient drifted target, as in ~\cite{deng2026generativemodelingdrifting}.
Let \(\mathrm{sg}(\cdot)\) denote stop-gradient. We define the target:
\begin{equation}
\widetilde{x}_i
=
\mathcal{C}_{c_i}\!\left(
\mathrm{sg}\!\left(\xgen_i + V_i\right)
\right).
\end{equation}
We then regress the generator output toward this target in full trajectory space.

\noindent \tbm{Weighted trajectory regression}
We use a weighted squared error that upweights action dimensions:
\begin{equation}
\min_\psi\; \mathbb{E}\left[\left\lVert \xgen_i - \widetilde{x}_i \right\rVert_W^2\right],
\end{equation}
with $\left\lVert u \right\rVert_W^2=
\sum_{t=0}^{H-1}
\left(
\lambda_s \left\lVert u[t,0{:}d_s]\right\rVert_2^2
+
\lambda_a \left\lVert u[t,d_s{:}D]\right\rVert_2^2
\right).$

In our implementation, this weighting is a practical guard against action-collapse: it biases the amortized refinement toward producing control-relevant diversity in the action block while still training the full window.
Algorithm~\ref{alg:kdp_train} summarizes one SGD step.

\begin{algorithm}[H]
\caption{Keyed drifting policy training (one SGD step)}
\label{alg:kdp_train}

\begin{mdframed}[
  backgroundcolor=AlgoBG,
  linecolor=AlgoBG, 
  linewidth=0pt,
  roundcorner=3pt,
  innertopmargin=6pt,
  innerbottommargin=6pt,
  innerleftmargin=8pt,
  innerrightmargin=8pt,
  skipabove=0pt,
  skipbelow=0pt
]
\begin{algorithmic}[1]
\Require Dataset minibatch \(\{\ypos_i\}_{i=1}^B\); key map \(\kkey\); generator \(g_\psi\);
temperatures \(\{\tau_m\}_{m=1}^M\); clamp operator \(\mathcal{C}\); clamp mask \(M\);
weights \(\lambda_a,\lambda_s\).

\State \(c_i \gets \kkey(\ypos_i)\), sample \(z_i \sim \mathcal{N}(0,I)\).
\State \(\xgen_i \gets \mathcal{C}_{c_i}\!\big(g_\psi(z_i,c_i)\big)\).
\State \(\yneg_j \gets \xgen_j\) and mask self-negatives (\(j=i\)) in repulsion weights.

\State \(V_i \gets 0\) for all \(i\).
\For{$m=1$ to $M$}
  \State \(\Wpos_{ij} \gets \mathrm{softmax}\!\big(-\dkey(\xgen_i,\ypos_j)/\tau_m\big)\).
  \State \(\Wneg_{ij} \gets \mathrm{softmax}\!\big(-\dkey(\xgen_i,\yneg_j)/\tau_m\big)\).
  \State \(\mupos_i \gets \sum_j \Wpos_{ij}\ypos_j,\quad \muneg_i \gets \sum_j \Wneg_{ij}\yneg_j\).
  \State \(V_i \gets V_i + \frac{1}{M}\big(\mupos_i - \muneg_i\big)\).
\EndFor
\State \(V_i \gets M \odot V_i\). // constraint-aware drift

\State \(\widetilde{x}_i \gets \mathcal{C}_{c_i}\!\big(\mathrm{sg}(\xgen_i + V_i)\big)\).
\State \(\Delta^s_{i,t} \gets \xgen_i[t,0{:}d_s]-\widetilde{x}_i[t,0{:}d_s]\)
\State \(\Delta^a_{i,t} \gets \xgen_i[t,d_s{:}D]-\widetilde{x}_i[t,d_s{:}D]\)
\State \(\mathcal{L} \gets \sum_{i,t}\Big(\lambda_s\lVert \Delta^s_{i,t}\rVert_2^2 + \lambda_a\lVert \Delta^a_{i,t}\rVert_2^2\Big)\)
\State Update \(\psi\) by SGD on \(\mathcal{L}\).
\end{algorithmic}
\end{mdframed}

\end{algorithm}

\subsection{Online receding-horizon planning}
\label{sec:method_planning}
At each environment step, KDP conditions on the current observation \(c\) and draws \(K\) samples in parallel:
\begin{equation}
\text{\footnotesize$\displaystyle
\hat{x}^{(k)} = \mathcal{C}_{c}\!\left(g_\psi(z_k,c)\right),
\qquad
z_k \sim \mathcal{N}(0,I),
\qquad
k=1,\dots,K.
$}
\end{equation}
A scorer \(J_\phi\) (a learned return model) selects a candidate,
\begin{equation}
\hat{x}^{*} = \arg\max_{k \in \{1,\dots,K\}} J_\phi(\hat{x}^{(k)},c),
\end{equation}
the executed action is the first action block of the window:
\begin{equation}
a \gets \mathrm{act}(\hat{x}^{*})[0] = \hat{x}^{*}[0,d_s{:}D].
\end{equation}

\begin{algorithm}[H]
\caption{Receding-horizon planning with KDP}
\label{alg:kdp_plan}

\begin{mdframed}[
  backgroundcolor=AlgoBG,
  linecolor=AlgoBG,
  linewidth=0pt,
  roundcorner=3pt,
  innertopmargin=6pt,
  innerbottommargin=6pt,
  innerleftmargin=8pt,
  innerrightmargin=8pt,
  skipabove=0pt,
  skipbelow=0pt
]
\begin{algorithmic}[1]
\Require Current context \(c\); generator \(g_\psi\); number of candidates \(K\); optional scorer \(J_\phi\).
\State Sample \(z_1,\dots,z_K \sim \mathcal{N}(0,I)\).
\State Generate candidates: \(\hat{x}^{(k)} \gets \mathcal{C}_c(g_\psi(z_k,c))\) for \(k=1,\dots,K\).
\If{$J_\phi$ is available}
    \State Select \(\hat{x}^{*} \gets \arg\max_k J_\phi(\hat{x}^{(k)},c)\).
\Else
    \State Select \(\hat{x}^{*} \gets \hat{x}^{(1)}\).
\EndIf
\State Execute \(a \gets \hat{x}^{*}[0,d_s{:}D]\) and repeat at the next control cycle.
\end{algorithmic}
\end{mdframed}

\end{algorithm}

This compute profile differs from diffusion planning: diffusion requires \(T\) {sequential} policy evaluations per environment step (each acting on a batch of \(K\) candidates under best-of-\(K\)), while KDP requires a single policy forward pass on \(K\) candidates (plus the optional scorer).
KDP therefore trades sequential refinement depth \(T\) for embarrassingly-parallel candidate count \(K\), which is easier to amortize on modern accelerators.

\section{Experiments}
\label{sec:experiments}

Our experiments are designed to validate four claims: (i) KDP supports amortized refinement, (ii) KDP offers a favorable compute--performance tradeoff relative to diffusion-based planners, (iii) KDP supports conditional planning, and (iv) the performance gains are attributable to specific components of the method, verified by ablations.

\subsection*{Experimental setup}
\label{sec:exp_setup}

\subsubsection{Simulated benchmarks}
\label{sec:exp_benchmarks}
We evaluate on D4RL benchmark \cite{fu_d4rl_2021}:
(i) locomotion,
(ii) goal-conditioned navigation,
(iii) long-horizon navigation,  
(iv) dexterous manipulation.
\begin{figure}[H]
    \centering
    \includegraphics[width=1\linewidth]{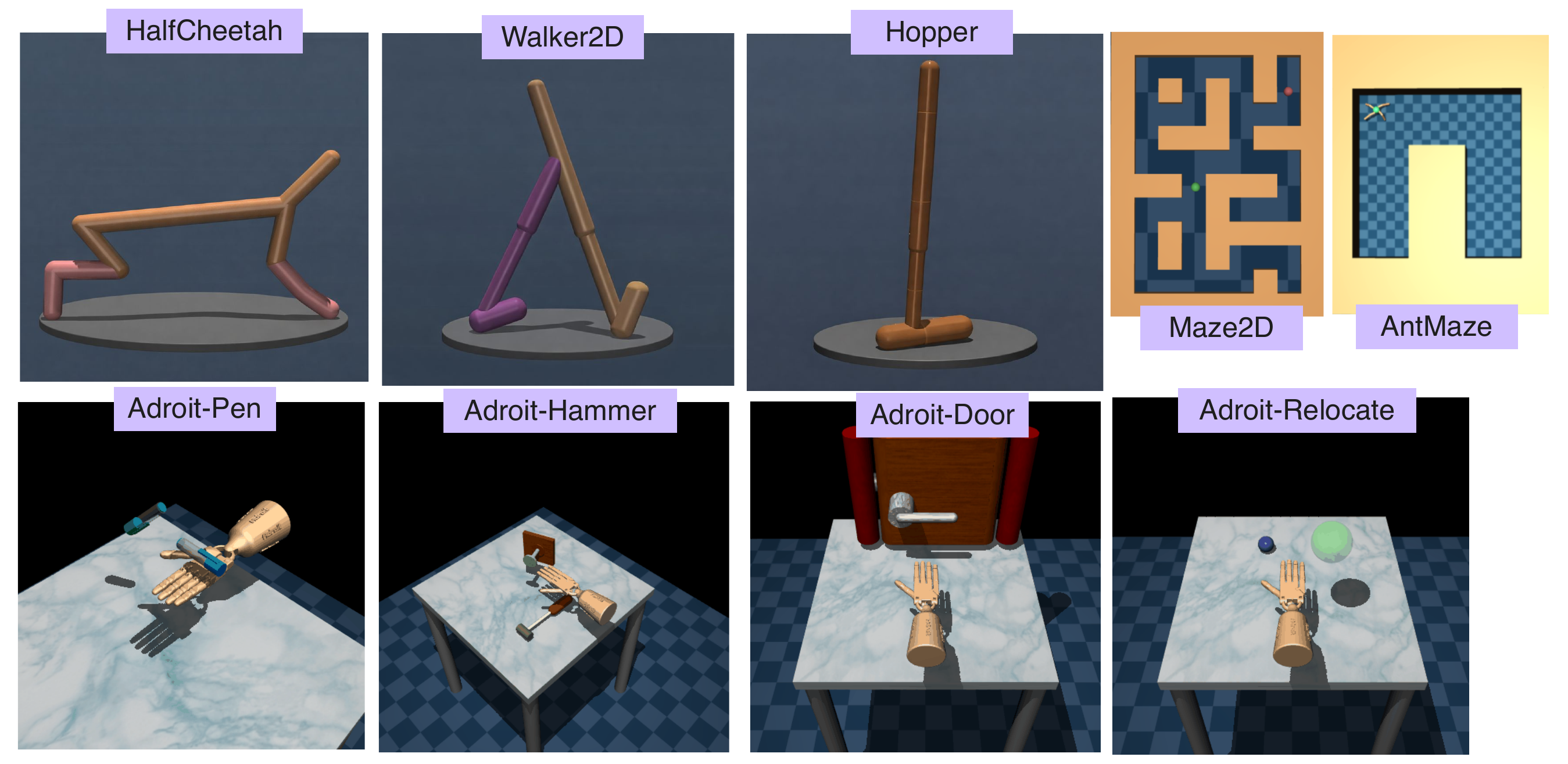}
    \caption{D4RL simulation environments.}
    \label{fig:placeholder}
\end{figure}

\subsubsection{Hardware implementation}
To showcase real-time performance of KDP, we evaluate on two domains: (i) quadrotor navigation and (ii) tabletop manipulation with a 6-DoF arm. 

\subsubsection{Metrics}
\label{sec:exp_planning}
We report the following per environment step:
(i) \textit{Sequential NFEs (NFE):} number of sequential policy forward calls required to produce candidates.
(ii) \textit{Batch-equivalent forwards (BEF):} total effective batch processed per step, accounting for candidate count.
(iii) \textit{Planning latency (ms/step) (PL):} end-to-end planning time per environment step.
(iv) \textit{End-to-end control latency (ms/step) (E2E):} planning latency plus environment and wrapper overhead.

\subsubsection{Baselines}
\label{sec:exp_baselines}
We separate baselines by their inference-time behavior: \textit{(i) fast-policy / offline-selection} baselines (BC \cite{bc}, CQL \cite{cql}, IQL \cite{iql}, Decision Transformer \cite{dt}, Trajectory Transformer \cite{tt}), and representative model-based offline RL methods (MOPO \cite{mopo}, MOReL\cite{morel}, and MBOP \cite{mbop}); \textit{(ii) Generative trajectory methods} (Diffuser \cite{janner_planning_2022}, trained on the same trajectory-window representation and evaluated with the same receding-horizon control as KDP).
For locomotion we additionally report recent low-NFE generative or trajectory-modeling approaches (DQL \cite{dql}, CAC \cite{cac}, and CBC \cite{cac}).  Where selection is used, we evaluate both (i) unranked sampling and (ii) ranked best-of-$K$ sampling using the same learned scorer used for KDP.

\subsection{Results: Short-horizon continuous control}
\label{sec:exp_locomotion}
We begin with D4RL locomotion, a hard benchmark for one-step generative planning. These tasks have dense rewards, relatively short horizons, and strong state-action regularities; as a result, they favor methods that can produce a sharp immediate action without needing explicit trajectory-level uncertainty. We therefore ask: can a one-step generative planner remain competitive while preserving multimodal sampling and collapsing inference to one NFE?

\begin{table}[H]
\centering
\begingroup
\scriptsize
\setlength{\tabcolsep}{0.5pt}
\renewcommand{\arraystretch}{1.18}

\arrayrulecolor{TblRule}
\setlength{\heavyrulewidth}{0.9pt}
\setlength{\lightrulewidth}{0.45pt}

\begin{tabularx}{\columnwidth}{@{}>{\hsize=1.9\hsize\linewidth=\hsize}L*{8}{>{\hsize=0.8\hsize\linewidth=\hsize}Y}@{}}
\toprule
\rowcolor{TblHeader}
& \multicolumn{2}{c}{\textbf{HalfCheetah}}
& \multicolumn{2}{c}{\textbf{Hopper}}
& \multicolumn{2}{c}{\textbf{Walker2D}}
& \textbf{Avg$\uparrow$} & \textbf{NFE$\downarrow$} \\
\rowcolor{TblHeader}
& \textcolor{black!70}{\scriptsize ME} & \textcolor{black!70}{\scriptsize M}
& \textcolor{black!70}{\scriptsize ME} & \textcolor{black!70}{\scriptsize M}
& \textcolor{black!70}{\scriptsize ME} & \textcolor{black!70}{\scriptsize M}
& & \\
\midrule

\sectionband{Offline selection}

BC \cite{bc}   & 55.2 & 42.6 & 52.5 & 52.9 & 107.5 & 75.3 & 64.3 & -- \\
CQL \cite{cql}  & 91.6 & 44.0 & 105.4 & 58.5 & 108.8 & 72.5 & 80.1 & -- \\
IQL \cite{iql}  & 86.7 & 47.4 & 91.5  & 66.3 & 109.6 & 78.3 & 80.0 & -- \\
DT \cite{dt}   & 86.8 & 42.6 & 107.6 & 67.6 & 108.1 & 74.0 & 81.1 & -- \\
TT \cite{tt}   & 95.0 & 46.9 & 110.0 & 61.1 & 101.9 & 79.0 & 82.3 & -- \\
MOPO \cite{mopo} & 63.3 & 42.3 & 23.7  & 28.0 & 44.6  & 17.8 & 36.6 & -- \\
MOReL\cite{morel} & 53.3 & 42.1 & 108.7 & \textbf{95.4} & 95.6 & 77.8 & 78.8 & -- \\
MBOP \cite{mbop} & \textbf{105.9} & 44.6 & 55.1 & 48.8 & 70.2 & 41.0 & 60.9 & -- \\

\sectionband{Generative trajectory planners}

DQL \cite{dql}     & 96.8 & \textbf{69.1} & \textbf{111.1} & 90.5 & 110.1 & 87.0 & \textbf{91.1} & 5 \\
CAC \cite{cac}     & 84.3 & \textbf{69.1} & 100.4 & 80.7 & \textbf{110.4} & 83.1 & 88.0 & 2 \\
CBC \cite{cac}     & 32.7 & 31.0 & 90.6 & 71.7 & \textbf{110.4} & 83.1 & 69.9 & 2 \\
Diffuser \cite{janner_planning_2022} & 88.9 & 42.8 & 103.3 & 74.3 & 106.9 & 79.6 & 82.6 & 20 \\

\rowcolor{ModernPink}
\textbf{\keyimg KDP}
& 92.5 & 62.1
& 103.6 & 90.3
& 108.5 & \textbf{87.2}
& 90.7 & \textbf{1} \\
\bottomrule
\end{tabularx}
\endgroup

\caption{Performance in short horizon control on D4RL Locomotion (ME=Medium-Expert, M=Medium).}
\label{tab:gym-mujoco_onecol}
\end{table}

The answer is yes (refer Table \ref{tab:gym-mujoco_onecol}). KDP is competitive with the strongest low-NFE generative trajectory methods while substantially improving over Diffuser. Locomotion is not the benchmark that most naturally favors trajectory sampling: when the correct action is strongly determined by the current state, conservative value-based methods and sequence models remain extremely difficult to beat. The key observation is that KDP does not need a diffusion-style denoising chain to remain in the same performance band as these methods. In other words, moving to one-step inference does not collapse the planner into a weak imitation policy. A second point is that the performance pattern is not uniform across dataset qualities. Medium-only datasets place greater pressure on the model to preserve diverse but dynamically consistent futures under weaker support, while medium-expert settings more strongly reward sharp action selection and better value discrimination. KDP remains robust across both regimes, which suggests that the gain is not coming from overfitting a single dataset type, but from replacing iterative refinement with a condition-aware amortized update that still preserves useful trajectory diversity.

\noindent \textit{\textbf{Ablation: }Compute--performance tradeoff.}
The systems-level distinction becomes much clearer when we look at inference cost (refer Table \ref{tab:locomotion_compute_a100_onecol}. Diffuser scales with the number of {sequential} denoising steps, so increasing candidate count compounds an already expensive inner loop. KDP changes that scaling law: candidate count grows primarily through parallel batch size, while the sequential depth of the policy remains fixed at one.

\begin{table}[H]
\centering
\begingroup
\scriptsize
\setlength{\tabcolsep}{1pt}
\renewcommand{\arraystretch}{1.18}

\arrayrulecolor{TblRule}
\setlength{\heavyrulewidth}{0.9pt}
\setlength{\lightrulewidth}{0.45pt}

\begin{tabularx}{\columnwidth}{@{}
  >{\raggedright\arraybackslash}p{0.30\columnwidth}
  Y Y Y Y @{}}
  \toprule
\rowcolor{TblHeader}
\textbf{Method}
& \textbf{NFE$\downarrow$}
& \textbf{BEF$\downarrow$}
& \textbf{PL}
& \textbf{E2E} \\
\midrule

BC \cite{bc} & \textbf{1} & \textbf{1} & \textbf{1.01} & \textbf{1.22} \\

D-UR($K{=}1,T{=}20$)  & 20 & 20   & 149.32 & 149.51 \\
D-R($K{=}16,T{=}20$) & 20 & 320  & 172.16 & 172.34 \\
D-R$K{=}64,T{=}20$) & 20 & 1280 & 279.98 & 280.03 \\

\rowcolor{ModernPink}
\keyimg\textbf{KDP-UR}($K{=}1$)  & \textbf{1} & \textbf{1}   & 1.98 & 2.11 \\
\rowcolor{ModernPink}
\keyimg\textbf{KDP-R}($K{=}16$) & \textbf{1} & 32  & 3.97 & 4.10 \\
\rowcolor{ModernPink}
\keyimg\textbf{KDP-R}($K{=}64$) & \textbf{1} & 128 & 4.76 & 4.96 \\
\bottomrule
\end{tabularx}
\endgroup
\caption{\normalfont Analysis of planning cost. {Hardware:} 1xNVIDIA A100. (D=Diffuser, UR=Unranked, R=Ranked)}
\label{tab:locomotion_compute_a100_onecol}
\end{table}

Ranked planning is only useful when extra candidates are cheap enough to sample inside the control budget. For Diffuser, larger $K$ becomes a latency bottleneck because every additional candidate inherits the full denoising stack. For KDP, increasing $K$ remains a practical knob: ranking acts as an inexpensive robustness improvement. 

\subsection{Results: Long-horizon, goal-conditioned planning}
\label{sec:exp_goal}
We next evaluate on Maze2D and AntMaze, where the core challenge is propagating conditioning information across a long horizon under delayed or sparse reward. These tasks stress the failure mode that motivated KDP.

\subsubsection{Maze2D: horizon and candidate scaling}
KDP improves on Diffuser while operating at essentially real-time planning cost (refer Table \ref{longhor}), which suggests that the gain is not just from being cheaper, but from using a more appropriate training signal for conditional trajectory generation. In Maze2D, once the neighborhood structure is aligned with the conditioned state/goal geometry, one-step generation appears sufficient to recover strong planning behavior. This result suggest that in a genuinely trajectory-level conditional problem, amortizing refinement into training is not merely a systems trick, it can also be the better inductive bias.

\begin{table}[H]
\centering
\begingroup
\scriptsize
\setlength{\tabcolsep}{0.1pt}
\begin{tabularx}{\columnwidth}{@{}>{\raggedright\arraybackslash}p{0.15\columnwidth}*{7}{Y}@{}}
\toprule
\rowcolor{TblHeader}
\textbf{Method}
& \multicolumn{4}{c}{\textbf{Maze2D}}
& \multicolumn{3}{c}{\textbf{AntMaze}} \\
\rowcolor{TblHeader}
  & \textbf{UM} & \textbf{Med.} & \textbf{Lar.} & \textbf{PL}
& \textbf{UM} & \textbf{Med.} & \textbf{PL} \\
\cmidrule(lr){2-5}\cmidrule(lr){6-8}
CQL \cite{cql}     & 5.7  & 5.0  & 12.5 & --  & \textbf{84.0} & 53.7 & -- \\
IQL \cite{iql}     & 47.4 & 34.9 & 58.6 & --  & 62.2 & \textbf{70.0} & -- \\
\midrule
Diffuser \cite{janner_planning_2022} & 113.9 & {121.5} & {123.0} & 3.848 & 76.0 & 31.9 & 6.153 \\
\rowcolor{ModernPink}
\keyimg\textbf{KDP}
         & \textbf{122.3} & \textbf{130.4} & \textbf{133.0} & \textbf{0.031}
         & 81.3 & 67.6 & \textbf{0.052} \\
\bottomrule

\end{tabularx}
\endgroup
\caption{Long horizon manipulation analysis. PL is in s/step. (UM=U-Maze, Med.=Medium, Lar.=Large)}
\label{longhor}
\end{table}

A second takeaway is that candidate scaling is only useful when the sampler stays inside the latency budget. Maze2D makes this visible because longer horizons and goal constraints make single-sample failure more likely, so best-of-$K$ selection should matter. KDP benefits from this regime precisely because it can afford larger candidate sets without paying a sequential-denoising penalty.

\subsubsection{AntMaze: long-horizon sparse-reward planning}
AntMaze is substantially harder than Maze2D because success depends not only on reaching the right goal region, but on discovering and maintaining globally correct route structure over long horizons under sparse feedback. 
The results reflect this difficulty. KDP retains its latency advantage and remains competitive with diffusion-based planning, but the gap over strong offline RL baselines narrows relative to Maze2D. AntMaze is a setting where more sampling steps alone are not enough: route selection, better goal representations, stronger ranking, or hierarchical structure matter more than simply denoising longer. KDP preserves fast, condition-aware trajectory generation even in this hard sparse-reward regime, and thus provides a stronger foundation for future improvements in selection or hierarchy.

\noindent \textit{\textbf{Ablation: }Training/objective.}
\label{sec:exp_train_ablate}
Table~\ref{tab:ablate_objective} isolates the training-time components of KDP. We ablate the neighborhood definition by replacing the default condition-keyed distance with full-window distances, test the role of collapse prevention by including self-negatives or removing repulsion entirely, and study the stability contribution of drift normalization and temperature averaging.
\begin{table}[H]
\scriptsize
\centering
\setlength{\tabcolsep}{1pt}
\renewcommand{\arraystretch}{1.18}

\arrayrulecolor{TblRule}
\setlength{\heavyrulewidth}{0.9pt}
\setlength{\lightrulewidth}{0.45pt}

\begin{tabularx}{\columnwidth}{@{}
  >{\raggedright\arraybackslash}p{0.46\columnwidth}
  Y Y Y @{}}\toprule
\rowcolor{TblHeader}
\textbf{Ablation} & \textbf{Locom.$\uparrow$} & \textbf{Maze2D$\uparrow$} & \textbf{Act. div.$\uparrow$} \\
\midrule
Full method (KDP)                      & \textbf{90.7}  & 128.2 & 0.93 \\
No keying (full-window distances)      & 2.91           & 9.6   & 0.14 \\
Include self-negatives                 & 78.08          & 91.3  & 0.93 \\
Attraction-only (no repulsion)         & 1.05           & 3.7   & 2.2e6 \\
No drift normalization                 & 86.93          & \textbf{131.5} & 0.93 \\
Single $\tau$                          & \textbf{92.37} & 102.1 & 0.93 \\
\bottomrule
\end{tabularx}
\caption{\normalfont Training objective ablations. (Locom.=Locomotion, Act. div.=Action diversity)}
\label{tab:ablate_objective}
\end{table}
The ablations show that the method depends first and foremost on {conditioning-aware neighborhoods}. Removing keying and computing similarity in full trajectory space causes performance to collapse, confirming that the central failure mode is not lack of model capacity but a misaligned notion of neighborhood under conditioning. Repulsion is also essential: when it is removed, performance collapses and the action-diversity statistic explodes, indicating uncontrolled drift rather than useful multimodality. Excluding self-negatives provides a smaller but still consistent gain, suggesting that once repulsion is present, its quality matters even if it is not the dominant factor. Drift normalization is less critical--performance remains strong without it and can even improve slightly on Maze2D--but normalization improves robustness across settings and makes the training signal better behaved overall. Finally, using a single temperature degrades Maze2D, which is consistent with our intuition that long-horizon goal-conditioned planning benefits from neighborhoods defined at multiple scales.

\subsection{Results: Dexterous manipulation and high-DoF control}
\label{sec:exp_manip}
We now turn to a much harder setting: dexterous manipulation, where small action errors are amplified by contacts, underactuation, and high-dimensional control.

\subsubsection{Adroit results}
The manipulation results make the benefit of KDP more pronounced than in locomotion. Unlike locomotion, where smooth dynamics often allow strong fast-policy baselines to dominate, Adroit requires the sampled plan to remain precise under contact and over multiple coupled degrees of freedom. In this setting, the one-step drift objective appears to align better with the control problem than iterative denoising of the entire trajectory window.
\begin{table}[H]
\centering

\scriptsize
\setlength{\tabcolsep}{0.5pt}
\renewcommand{\arraystretch}{1.18}

\arrayrulecolor{TblRule}
\setlength{\heavyrulewidth}{0.9pt}
\setlength{\lightrulewidth}{0.45pt}

\begin{tabularx}{\columnwidth}{@{}l *{4}{Y} @{}}
\toprule
\rowcolor{TblHeader}
\textbf{Task}
& \multicolumn{2}{c}{\textbf{Diffuser \cite{janner_planning_2022}}}
& \multicolumn{2}{c}{\keyimg\textbf{KDP}} \\
\rowcolor{TblHeader}
& \textcolor{black!70}{\scriptsize Score} & \textcolor{black!70}{\scriptsize T(s)}
& \textcolor{black!70}{\scriptsize Score} & \textcolor{black!70}{\scriptsize T(s)} \\
\midrule

pen-clone      & 10.7 & 1.634 & \cellcolor{ModernPink}\textbf{53.4} & \cellcolor{ModernPink}\textbf{0.020} \\
door-clone     & 56.7 & 1.598 & \cellcolor{ModernPink}\textbf{61.8} & \cellcolor{ModernPink}\textbf{0.024} \\
hammer-clone   & 53.1 & 1.532 & \cellcolor{ModernPink}\textbf{79.6} & \cellcolor{ModernPink}\textbf{0.034} \\
relocate-clone & 56.2 & 1.685 & \cellcolor{ModernPink}\textbf{62.8} & \cellcolor{ModernPink}\textbf{0.036} \\
& & & & \\
\bottomrule
\end{tabularx}

\caption{\normalfont Adroit results.}
\label{tab:adroit_main}

\end{table}
Two insights matter here (refer Table \ref{tab:adroit_main}). First, the gain is not just faster than Diffuser, it is faster {and} better on tasks where poor action structure is immediately punished. Second, the improvement is consistent with the way KDP is trained: the action block is explicitly emphasized in the regression objective, so the model is encouraged to learn plans whose control-relevant portion is sharp and usable at execution time. In high-DoF manipulation, that appears to be a more favorable bias than repeatedly refining the full window online.
These results are also important for deployment---planning delays on the order of seconds are often unacceptable regardless of final normalized score, because contact transitions and object motion evolve on much faster timescales.

\noindent \textit{\textbf{Ablation:} Action chunking.}
\label{sec:exp_action_chunk}
Action chunking provides a second lens on trajectory quality (refer Table \ref{tab:action_chunk_adroit_all}). If a sampled window is only trustworthy at its very first action, then executing multiple actions before replanning should quickly degrade performance. If, instead, the sampled trajectory is locally coherent, modest chunk sizes should improve stability by filtering high-frequency planning noise and reducing replanning overhead. The chunking trends support the latter interpretation for KDP. Moderate chunk sizes improve performance, which suggests that KDP's windows are not merely optimized for the first action token. This is the behavior one would want before moving to hardware: the planner should not collapse when sensing, actuation, and planning are slightly out of sync. 
\begin{table}[H]
\centering

\scriptsize
\setlength{\tabcolsep}{0.5pt}
\renewcommand{\arraystretch}{1.18}

\arrayrulecolor{TblRule}
\setlength{\heavyrulewidth}{0.9pt}
\setlength{\lightrulewidth}{0.45pt}

\begin{tabularx}{\columnwidth}{@{}
  >{\raggedright\arraybackslash}p{0.40\linewidth}
  >{\centering\arraybackslash}p{0.03\linewidth}
  Y Y @{}}
\toprule
\rowcolor{TblHeader}
\textbf{Task} & \textbf{$L$} & \textbf{Diff.} & \keyimg\textbf{KDP} \\
\midrule
adroit-pen      & 1 & 38\% & \cellcolor{ModernPink} 57\% \\
adroit-pen      & 8 & 41\% & \cellcolor{ModernPink} 73\% \\
adroit-hammer   & 1 & 47\% & \cellcolor{ModernPink} 62\% \\
adroit-hammer   & 8 & 41\% & \cellcolor{ModernPink} 63\% \\
adroit-relocate & 1 & 35\% & \cellcolor{ModernPink} 65\% \\
adroit-relocate & 8 & 64\% & \cellcolor{ModernPink} 71\% \\
\bottomrule
\end{tabularx}

\caption{\normalfont Action chunking.}
\label{tab:action_chunk_adroit_all}

\end{table}

\subsection{Real-world hardware experiments}
\label{sec:exp_hardware}
\begin{figure}[H]
    \centering
    \includegraphics[width=0.9\linewidth]{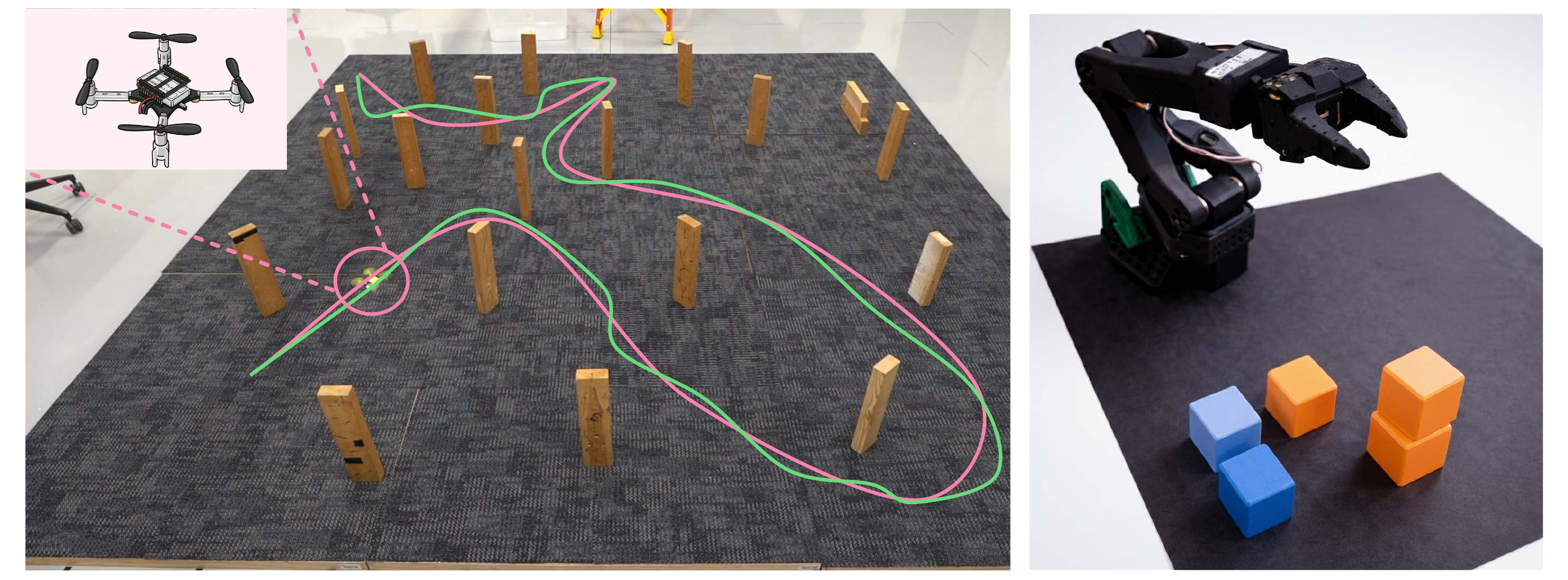}
    \caption{Qualitative real-time hardware results. Left: Crazyflie navigation trajectories for KDP (pink) and Diffuser (green). Right: SO-100 manipulation; the full behavior is best seen in the supplementary.}
    \label{fig:hw_drift}
\end{figure}
We finally evaluate KDP on two real-world domains. We consider two representative settings with different failure modes: (i) \textit{navigation}, using a Crazyflie~2.0 micro-quadrotor to fly from start to goal through cluttered indoor space, and (ii) \textit{manipulation}, using an SO-100 arm for tabletop pick-place and stacking tasks.

For both domains, we report task success (Succ.), median time-to-success over successful trials (TTS), the achieved replanning frequency and time of the outer closed-loop controller (Replan, E2E), and measured planner latency per control cycle (Planner).

\begin{table}[H]
\scriptsize
\centering

\setlength{\tabcolsep}{1.0pt}
\renewcommand{\arraystretch}{1.18}

\arrayrulecolor{TblRule}
\setlength{\heavyrulewidth}{0.9pt}
\setlength{\lightrulewidth}{0.45pt}

\begin{tabularx}{\columnwidth}{@{}l Y Y Y Y Y Y@{}}
\toprule
\rowcolor{TblHeader}
\textbf{Domain} & \textbf{Method} & \textbf{Succ. (\%)} & \textbf{TTS (s)} & \textbf{Replan (Hz)} & \textbf{Planner p50} & \textbf{E2E p50} \\
\midrule

Navigation
& Diffuser  & 92 & 12.5 & 4.1   & 242 & 312 \\
Navigation
& \cellcolor{ModernPink}\keyimg\textbf{KDP} 
& \cellcolor{ModernPink}94 & \cellcolor{ModernPink}12.0 & \cellcolor{ModernPink}38
& \cellcolor{ModernPink}8 & \cellcolor{ModernPink}26 \\

\addlinespace[2pt]
Manipulation
& Diffuser  & 88 & 35 & 3   & 311 & 410 \\
Manipulation
& \cellcolor{ModernPink}\keyimg\textbf{KDP} 
& \cellcolor{ModernPink}90 & \cellcolor{ModernPink}32 & \cellcolor{ModernPink}17
& \cellcolor{ModernPink}9 & \cellcolor{ModernPink}47 \\
\bottomrule
\end{tabularx}

\caption{\normalfont {Real-world closed-loop results on a Crazyflie and an SO-100 arm.}
{Setup:} Diffuser ($K{=}64$, $T{=}20$); KDP ($K{=}64$).}
\label{tab:realworld_closedloop}
\end{table}

The advantage of KDP is not merely that it preserves task success while using fewer NFEs, but that it moves generative planning into a qualitatively different operating regime. On both platforms, KDP matches or slightly improves closed-loop task performance while sustaining substantially higher replanning rates and dramatically lower control-cycle latency. This matters because real-world failures are often caused less by the nominal quality of a single plan than by the fact that the plan becomes stale before it can be refreshed. The results therefore validate our central claim that amortizing refinement into training retains the benefits of trajectory-level generative planning while making the planner reactive enough for real-time deployment.




\section{Conclusion}
\label{sec:conclusion}

We introduced \keyimg{\modpink{Keyed Drifting Policies (KDP)}}, a one-step conditional trajectory generator trained with a keyed drift field. The central design choice is \emph{conditioning-aware neighborhoods}: drift weights are computed in a compact key space aligned with the condition, while drift updates act in full trajectory space. Combined with attraction--repulsion updates, self-negative masking, constraint masking, and a stop-gradient drifted target, KDP avoids the action-collapse failure mode observed under naïve full-window similarity.
Empirically, KDP achieves competitive or improved performance on standard offline RL planning benchmarks while dramatically reducing planning latency relative to diffusion sampling. Beyond simulation, we demonstrated real-time deployment on hardware in navigation and manipulation settings, where KDP sustains substantially higher replanning rates with comparable task success.
\textit{Limitations:} goal-conditioned and sparse-reward domains remain challenging. 

\bibliographystyle{plain}
\bibliography{references}
\end{document}